\pdfoutput=1

\documentclass[11pt]{article}

\usepackage[]{acl}

\usepackage{times}
\usepackage{latexsym}

\usepackage[T1]{fontenc}

\usepackage[utf8]{inputenc}

\usepackage{microtype}

\usepackage{multirow}
\usepackage{booktabs}
\usepackage{graphicx}
\usepackage{subfigure}
\usepackage{amssymb}
\usepackage{CJKutf8}
\usepackage{amsmath} 
\usepackage{hyperref}
\usepackage{enumitem}
\usepackage{color}
\usepackage{algorithm}
\usepackage{algpseudocode}

\title{Modeling Structural Similarities between Documents for Coherence Assessment with Graph Convolutional Networks}

\author{Wei Liu$^{1}$, Xiyan Fu$^{2}$, Michael Strube$^{1}$ \\ $^{1}$Heidelberg Institute for Theoretical Studies gGmbH  \\ $^{2}$Institute for Computational Linguistics, 
Heidelberg University \\ \texttt{wei.liu@h-its.org}, \texttt{fu@cl.uni-heidelberg.de}, \\ \texttt{michael.strube@h-its.org}}

\begin{document}
\maketitle
\maketitle

\begin{abstract}
Coherence is an important aspect of text quality, and various approaches have been applied to coherence modeling. However, existing methods solely focus on a single document's coherence patterns, ignoring the underlying correla\-tion between documents. We investigate a GCN-based coherence model that is capable of capturing structural similarities between documents. Our model first creates a graph structure for each document, from where we mine different subgraph patterns. We then construct a hetero\-geneous graph for the training corpus, connecting documents based on their shared subgraphs. Finally, a GCN is applied to the heterogeneous graph to model the connectivity relationships. We evaluate our method on two tasks, assessing discourse coherence and automated essay scoring. Results show that our GCN-based model outperforms all baselines, achieving a new state-of-the-art on both tasks.

\end{abstract}

\section{Introduction}
Coherence describes the relationship between sentences that makes a group of sentences logically connected rather than just a random collection of them \cite{speech}. It is an important aspect of text quality \citep{mcnamara2010linguistic}, and its modeling has been applied in many downstream tasks, including summarization \citep{parveen-etal-2015-topical, wu2018learning}, dialogue generation \citep{mesgar-etal-2020-dialogue,xu-etal-2021-discovering}, machine translation \citep{xiong2019modeling,tan-etal-2019-hierarchical} and document-level text generation \citep{wang-etal-2021-building,diao-etal-2021-tilgan}. Given the importance of the task, there is a long line of methods proposed for coherence modeling.

Previous models leverage linguistic features to solve the problem. For example, entity grid-based methods \citep{barzilay-lapata-2005-modeling, elsner-charniak-2011-extending} capture the entity transition between adjacent sentences of a text to model local coherence; in contrast, graph-based models \citep{guinaudeau-strube-2013-graph,mesgar-strube-2015-graph} measure coherence using the entity-graph of a document. Recently, neural network models \citep{li-hovy-2014-model,li-jurafsky-2017-neural,mesgar-strube-2018-neural,xu-etal-2019-cross,farag-yannakoudakis-2019-multi,moon-etal-2019-unified,jeon-strube-2020-centering,jeon-strube-2020-incremental,mesgar-etal-2021-neural-graph} have been applied to the task due to their strength in representation learning and feature combination. Those models learn a document's representation from word embeddings or pre-trained language models, giving significantly better performance than previous statistical methods.

\begin{figure}[t]
\centering\includegraphics[scale=0.55,trim=0 0 0 0]{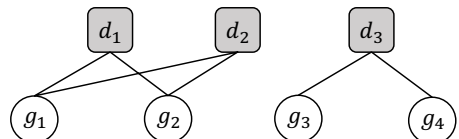}
\setlength{\abovecaptionskip}{1pt} 
\caption{An example of structurally similar documents connected through subgraphs, in which $d_2$ is more similar in structure to $d_1$ than $d_3$. $d_i$ denotes the $i$-th document and $g_j$ denotes the $j$-th subgraph pattern, an edge will be built between them if the sentence graph of $d_i$ contains subgraph $g_j$. }
\label{fig-motivation}
\vspace{-8pt}
\end{figure}

However, one drawback of existing neural-based methods is that they solely focus on extracting features within a single document, ignoring the underlying correlations between documents. Coherence describes how sentences of a text connect to each other ~\citep{tanya1980,foltz1998measurement,schwarz2001establishing}. Theoretically, documents with similar connection structures should tend to have similar degrees of coherence, which can be useful prior knowledge for coherence modeling. For example, a model is more likely to accurately assess a new document's coherence if it can refer to the labels of known documents with a similar organizational structure (see Appendix \ref{app.sub.text_exam} for an example).

To fill this gap, we investigate a graph-based approach to model the correlation between documents from the perspective of structural similarity. The main idea is to connect structurally similar documents through a graph and capture those connectivity relationships using Graph Convolutional Networks (GCN) \citep{kipf2017semi}. In particular, inspired by \citet{guinaudeau-strube-2013-graph}, we first represent a document as a sentence graph, where nodes are sentences and two nodes will be connected if they contain semantically related nouns. Our method further converts each sentence graph into a subgraph set as it proves to be an efficient approach for measuring the topological similarity between graphs \citep{shervashidze2009efficient,kondor2009graphlet}. Then, we construct a heterogeneous graph for the training corpus, containing document and subgraph nodes, based on subgraphs shared between documents. In this way, structurally-similar documents are explicitly linked through the subgraphs (shown in Figure \ref{fig-motivation}). Finally, a GCN is applied to the heterogeneous graph to learn the representation of document nodes while considering the connections between them.

We evaluate our method on two benchmark tasks\footnote{\url{https://github.com/liuwei1206/StruSim}}: assessing discourse coherence and automatic essay scoring. Experimental results show that our method significantly outperforms a baseline model that does not consider structural similarities between documents, achieving a new state-of-the-art performance on both tasks. In addition, we provide a comprehensive comparison and detailed analysis, which empirically confirm that structural similarity information helps to mitigate the effects of uneven label distributions in datasets and improve the model’s robustness across documents with different lengths.

\section{Related Work}
Our work is related to text coherence modeling and graph neural networks (GNN)-based methods for natural language processing (NLP).

\noindent{\textbf{Coherence Modeling}.} Inspired by Centering Theory ~\cite{grosz-etal-1995-centering}, ~\citet{barzilay-lapata-2005-modeling,barzilay-lapata-2008-modeling} propose an entity-based approach (the entity grid) to assess coherence by considering entity transitions between adjacent sentences of a text. The entity grid model has been improved by grouping entities based on their semantic relatedness ~\citep{filippova-strube-2007-extending}, incorporating entity-specific features ~\citep{elsner-charniak-2011-extending}, replacing grammatical roles with discourse roles ~\citep{lin-etal-2011-automatically}. On the other hand, ~\citet{guinaudeau-strube-2013-graph} propose an entity graph to capture entity transitions between not only adjacent sentences but also non-adjacent ones. Motivated by the functional sentence perspective of text coherence ~\citep{danes1974functional}, ~\citet{mesgar-strube-2015-graph,mesgar-strube-2016-lexical} improve the entity graph with graph-based features extracted from text structures. Similarly, we also leverage the structural features of texts. However, instead of feeding individual documents' structure into the model as coherence patterns, we use them to capture the underlying correlation between documents.

\begin{figure*}[t]
\centering\includegraphics[scale=0.54,trim=0 0 0 0]{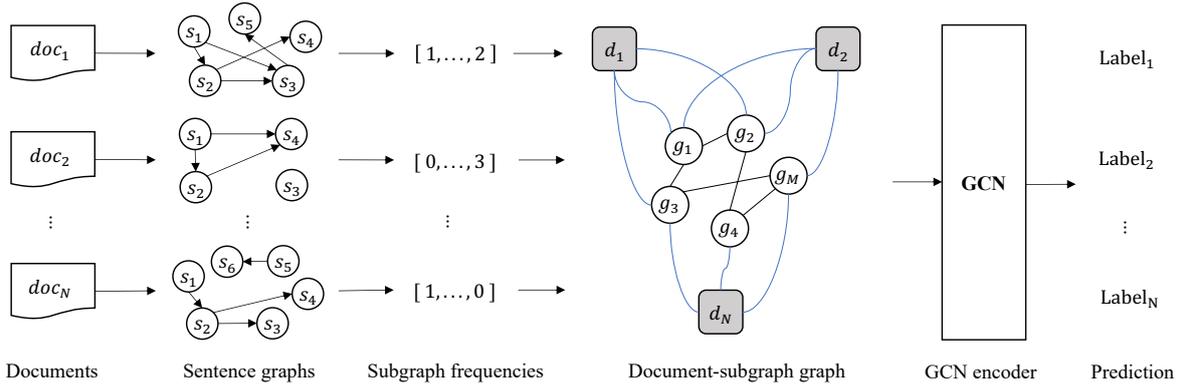}
\setlength{\abovecaptionskip}{8pt} 
\caption{Overview of the proposed approach. Our method identifies a document's graph structure, converts the graph into a subgraph set, constructs a corpus-level graph based on the shared subgraphs between structurally similar documents, and finally encodes those connections with a GCN. For simplicity, we only show three documents and five subgraphs and limit the number of sentences in a document. $s_u$, $d_i$, and $g_j$ denote the $u$-th sentence in a document, the $i$-th document in the training corpus, and the $j$-th defined subgraph, respectively. }
\label{fig-framework}
\vspace{-8pt}
\end{figure*}

With the advent of deep learning, neural networks have been applied to coherence modeling. \citet{li-hovy-2014-model,xu-etal-2019-cross} learn to assess coherence by training a model to distinguish coherent texts from incoherent ones using different neural encoders. ~\citet{tien-nguyen-joty-2017-neural,joty-etal-2018-coherence} extend the entity grid model with a convolutional neural network.  ~\citet{moon-etal-2019-unified} propose to enrich the coherence features of a document by considering discourse relations and syntactic patterns within it. ~\citet{jeon-strube-2020-centering} design structure-guided inter-sentence attention to learn a document's local and global coherence patterns. Inspired by human reading habits,  ~\citet{jeon-strube-2020-incremental} investigate a model to measure a document's coherence by incrementally interpreting sentences. Our work is in line with the above approaches to learning a coherence model based on neural networks. The main difference is that the above neural models focus on extracting features within a single document, whereas our graph-based approach aims to study the effectiveness of correlations between documents. Specifically, rooting on the linguistic definition of text coherence, we model the correlation from the perspective of structural similarities between documents.

\noindent{\textbf{GNN-based Methods for NLP}.} Graph neural networks are a family of neural networks that operate naturally on graphs. Many NLP problems can be expressed with a graph structure, so there is a surge of interest in applying GNNs for NLP tasks. ~\citet{marcheggiani-titov-2017-encoding} present a Syntactic GCN to learn latent feature representations of words in a sentence over dependency trees for Semantic Role Labeling.  ~\citet{yasunaga-etal-2017-graph} propose a GCN-based multi-document summarization system that exploits the sentence relation information encoded in graph representations of document clusters. ~\citet{yao2019graph} build a graph containing documents and words as nodes and used the Text GCN to learn embeddings of words and documents. ~\citet{lv2020graph} design a graph-based approach to encode structural information from ConceptNet for commonsense question answering. Compared with existing work, our graph-based method is different in both motivation and graph construction. For example, we specially design subgraph nodes to connect documents with a similar structure for capturing the structural correlations between samples.

\section{Method}
Figure \ref{fig-framework} shows an overview of our proposed method. We describe step-by-step how to capture the structural similarities between documents, including i) identifying the structure of a document (Section \ref{sent-graph}); ii) representing the sentence graph of a document as a subgraph set (Section \ref{subgraph-set}); iii) building a corpus-level heterogeneous graph to connect structurally similar documents based on the shared subgraphs (Section \ref{doc-sub-graph}); iv) applying a GCN encoder to capture connectivity relationships between document nodes (Section \ref{gcn}).

\subsection{Sentence Graph}\label{sent-graph}
To model the structural similarities between documents, we need to identify each document's structure. We follow ~\citet{guinaudeau-strube-2013-graph} to represent a document as a directed sentence graph but with some modifications in graph construction. Specifically, in our implementation, two sentences are semantically connected if there are strong semantic relations between nouns in the two sentences. We use nouns instead of entities ~\citep{guinaudeau-strube-2013-graph} because the former shows better performance than the latter in modeling semantic connection between sentences ~\citep{elsner-charniak-2011-extending,tien-nguyen-joty-2017-neural}. 

Given a document, we use the Stanza toolkit ~\citep{qi-etal-2020-stanza} to segment it into sentences $\{s_1, s_2,...,s_L\}$ and recognize all nouns in each sentence. For a pair of sentences $s_u$ and $s_v$ ($u < v$), we compute the similarity score for each pair of nouns from them (one noun from $s_u$ and the other from $s_v$) and use the maximum similarity score to measure their semantic connection. The score between two nouns is obtained by calculating the cosine value of their embedding. If the maximum similarity score is greater than the preset threshold $\delta$, then the two sentences are considered semantically connected, and we add a directed edge between them (from $s_u$ to $s_v$). After computing all combinations of $s_u$ and $s_v$ ($u < v$) in the document, we can build a directed graph containing sentences as nodes (refer to Algorithm \ref{alg_sentgraph} in Appendix \ref{sec:al_graph}).

\subsection{Subgraph Set}\label{subgraph-set}
After obtaining the graph structure of documents, we represent each sentence graph as a subgraph set. The subgraph set is an efficient way to compare topological similarities between graphs \citep{shervashidze2009efficient}, which we can employ to compare documents in terms of structure.

\begin{figure}[t]
\centering\includegraphics[scale=0.55,trim=0 0 0 0]{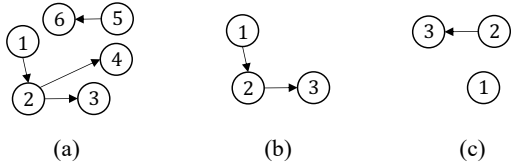}
\setlength{\abovecaptionskip}{1pt} 
\caption{An example of subgraphs, in which graph (b) and graph (c) are 3-node subgraphs of graph (a).}
\label{subgraph}
\vspace{-8pt}
\end{figure}

Graph $g$ is a subgraph of graph $G$ if the nodes in $g$ can be mapped to the nodes in $G$ and the connection relations within the two sets of nodes are the same. If the subgraph contains k nodes, we call it a $k$-node subgraph. In our method, we only consider subgraphs without backward edges. This is because when constructing the sentence graph, we process the document from left to right and never look back. We use weakly connected and disconnected subgraphs (shown in Figure \ref{subgraph}) since we empirically find they both reflect the properties of a document in terms of coherence.

Given a sentence graph $G_i$ of a document $d_i$, we first mine the contained $k$-node subgraphs by enumerating all combinations of $k$ nodes and corresponding edges in $G_i$. Subgraphs with inter-sentence distances greater than $w$ are filtered out because far-distant sentences are less likely to be related. It also reduces the search space when mining subgraphs. In the retained subgraphs, two can have the same structure but only differ in node IDs. We consider them as the same subgraph since they are isomorphic in graph theory. Then, we count the frequency of each k-node subgraph and identify the isomorphic subgraphs using the pynauty tool. Consequently, a sentence graph is represented as a k-node subgraph set  (refer to Algorithm \ref{alg_subgraph} in Appendix \ref{sec:al_graph}).

\subsection{Doc-subgraph Graph}\label{doc-sub-graph}
A graph is an efficient way to model the correlation between items and has been used in different domains, such as knowledge graphs ~\citep{carlson2010toward} and social networks ~\citep{tang2009relational}. We build a corpus-level undirected graph (on the training dataset), named doc-subgraph graph, to explicitly connect structurally similar documents through their shared subgraphs (shown in Figure \ref{fig-framework}). The graph contains document nodes and subgraph nodes, and the total number of nodes is the sum of the number of documents ($N$) and the number of k-node subgraph types ($M$) mined in Section \ref{subgraph-set}. We design two types of edges in the graph, including edges between document and subgraph, and edges between subgraphs. We build the first type of edge if a document's subgraph set contains a subgraph, and set its weight as the product of the subgraph's normalized frequency in the subgraph set and the subgraph's inverse document frequency in the corpus. The definition of inverse document frequency is adopted from TF-IDF, but here it represents how common a subgraph is across subgraph sets of all documents. As for the second kind of edge, we construct it between two subgraphs that appear in the same subgraph set of a document, and its weight is the co-occurrence probability of these two subgraphs. We model the co-occurrence information between subgraphs because it has been shown helpful for comparing similar structures between graphs ~\citep{kondor2009graphlet}.

Formally, we denote documents in a training corpus as $\mathbf{D}=\{d_1,d_2,...,d_N\}$ and all types of k-node subgraphs mined from the corpus as $\mathbf{SubG}=\{g_1, g_2, ..., g_M\}$. We use $G_i$ to denote the sentence graph of document $d_i$ and $F_i = \{f_{i1}, f_{i2},...,f_{iM}\}$ to denote the k-node subgraph set mined from $G_i$, where $f_{ij}$ denotes the frequency of subgraph $g_j$. We represent nodes in the doc-subgraph graph as $\mathbf{V} = \{v_1, ..., v_N, v_{N+1}, ..., v_{N+M}\}$, in which $\{v_1, ..., v_N\}$ are documents $\mathbf{D}$ and $\{v_{N+1},...,v_{N+M}\}$ are k-node subgraphs $\mathbf{SubG}$.

For any pair of document node $v_i\;(i \leq N)$ and subgraph node $v_{N+j} \;(j \leq M)$, we build an edge between them if $g_j$ appears in the subgraph set of $d_i$, i.e. $f_{ij} > 0$, and define the edge's weight as:
\begin{equation}
    \label{tf-idf}
    A_{i,N+j} = {\frac{f_{ij}}{\sum_{j'=1}^M f_{ij'}} \cdot \log\frac{N}{|d \in \mathbf{D} : g_j \in d |}} 
\end{equation}
where the first term is the normalized frequency of subgraph $g_j$ in subgraph set $F_i$, and the second term is an inverse document frequency factor, which diminishes the weight of subgraphs that occur frequently in subgraph sets and increases the weight of subgraphs that occur rarely. $|d \in \mathbf{D} : g_j \in d |$ represents the number of documents whose subgraph set contains subgraph $g_j$. $A$ denotes the adjacency matrix of the doc-subgraph graph with shape $(N+M) \times (N+M)$ and is initialized as zero matrix. To make the graph symmetrical, we set the value of $A_{N+j, i}$ to be the same as $A_{i,N+j}$. 

We also construct edges between any pair of subgraph nodes $v_{N+j}$ and $v_{N+j'} \;(j \leq M, j' \leq M, j \neq j')$ if $g_j$ and $g_{j'}$ co-occur in the subgraph set of a document, i.e. $ \exists \; d_i \in \mathbf{D}: f_{ij} > 0, \; f_{ij'} > 0 $. The weight is set as the Pointwise Mutual Information (PMI) of these two subgraphs, which is a popular way ~\citep{ghazvininejad-etal-2016-generating,yao2019graph} to measure co-occurrence information:
\begin{equation}
    A_{N+j,N+j'} = \log \frac{p(j,j')}{p(j)\;p(j')}
\end{equation}
\begin{equation}
    \begin{aligned}
        p(j) &= \frac{|d \in \mathbf{D}: g_j \in d|}{N} \\
        p(j,j') &= \frac{|d \in \mathbf{D}: g_j \in d, \; g_{j'} \in d|}{N} 
    \end{aligned}
\end{equation}

\noindent The PMI can be positive or negative, we follow previous work to clip negative PMI at 0 since this strategy works well across many tasks ~\citep{kiela-clark-2014-systematic,milajevs-etal-2016-robust,salle2019down}. 

\subsection{GCN Encoder}\label{gcn}
We adopt a GCN ~\citep{kipf2017semi} to encode the built doc-subgraph graph. GCN is a graph neural network which directly operates on graph-structured data. By integrating the normalized adjacency matrix, the GCN learns node representations based on both connectivity patterns and feature attributes of the graph ~\citep{li2018deeper}.

Formally, given the built graph with $(N+M)$ nodes, we represent the graph with an $(N+M) \times (N+M)$ adjacency matrix $A$. We first follow ~\citet{kipf2017semi} to add self-connections for each node:
\begin{equation}
    \tilde{A} = A + I_{N+M}
\end{equation}
where $I_{N+M}$ is an identity matrix. Then, a two-layer GCN is applied on the graph, and the convolution computation at the $l$-th layer is defined as:
\begin{equation}
    H^{(l)} = \sigma\left(\tilde{D}^{-\frac{1}{2}}\tilde{A}\tilde{D}^{-\frac{1}{2}}H^{(l-1)}{\rm \mathbf{W}}^{(l-1)}\right)
\end{equation}
Here, $\tilde{D}$ is the degree matrix (i.e. $\tilde{D}_{i,i}=\sum_j\tilde{A}_{i,j}$) and ${\rm \mathbf{W}}^{(l-1)}$ is a layer-specific trainable weight matrix. $\sigma$ is an activation function, such as ${\rm ReLU}$. $H^{(l)}$ is the output of $l$-th GCN layer; $H^{(0)}=X$, which is a matrix of node features. We use representations from the pre-trained model as features of document nodes due to its excellent performance on document-level tasks ~\citep{guo-nguyen-2020-document,yin-etal-2021-docnli,zhou2021document}. For subgraph nodes, since they have no textual contents, we set their features to zero vectors, which is a common setting in heterogeneous graphs \citep{jiheter}. Finally, we feed the outputs of the two-layer GCN into a softmax classifier:
\begin{equation}
    P = {\rm softmax}(H^{(2)})
\end{equation}
and train the model by minimizing the Cross-Entropy loss over document nodes:
\begin{equation}
    \mathcal{L} = -\sum_{i=1}^{N}\sum_{c=1}^{C}Y_{i,c} \cdot {\rm log}\left(P_{i,c}\right)
\end{equation}
where $Y_i$ is the label of document node $v_i$ with a one-hot scheme, $C$ is the number of classes.

While evaluating, for each document in the test corpus, we add it to the doc-subgraph graph, normalize the adjacent matrix of the new graph, and predict its label (refer to Appendix \ref{sec:train_and_eval}).

\section{Experiments}
\label{main-exp}
\subsection{Datasets}
\label{sec-dataset}
We evaluate the proposed method on two benchmark tasks, assessing discourse coherence (ADC) and automated essay scoring (AES). The descriptive statistics of the dataset for each task are shown in Appendix \ref{sec:dd_dataset}.

\begin{table*}[t]
\centering
\setlength{\tabcolsep}{4.8mm}
\scalebox{0.82}{
\begin{tabular}{l|cccc|c}
\hline
Model  & Yahoo\hspace{1.5em} & Clinton\hspace{1.4em} & Enron\hspace{1.5em} & Yelp\hspace{1.5em} & Avg \\ \hline
\citet{li-jurafsky-2017-neural} &    53.50\textsubscript{\; \; \; \,}   &     61.00\textsubscript{\; \; \; \,}    &    54.40\textsubscript{\; \; \; \,}   &   49.10\textsubscript{\; \; \; \,}   &  54.50   \\
\citet{lai-tetreault-2018-discourse} &    54.90\textsubscript{\; \; \; \,}   &    60.20\textsubscript{\; \; \; \,}     &   53.20\textsubscript{\; \; \; \,}    &   54.40\textsubscript{\; \; \; \,}   &  55.70   \\
\citet{mesgar-strube-2018-neural} &    47.30\textsubscript{\; \; \; \,}   &     57.70\textsubscript{\; \; \; \,}    &    50.60\textsubscript{\; \; \; \,}   &    54.60\textsubscript{\; \; \; \,}  &  52.55   \\
\citet{mesgar-strube-2016-lexical}\textsuperscript{${\dagger}$} & 61.30\textsubscript{0.84}   &    64.60\textsubscript{0.89}     &   55.74\textsubscript{0.90}    &   56.70\textsubscript{0.78}   &  59.59   \\
\citet{moon-etal-2019-unified}\textsuperscript{${\dagger}$} & 56.80\textsubscript{0.95}   &    60.65\textsubscript{0.76}     &   54.10\textsubscript{0.89}    &   55.85\textsubscript{0.85}   &  56.85   \\
\citet{jeon-strube-2020-centering}\textsuperscript{${\dagger}$} & 56.75\textsubscript{0.83}   &    62.15\textsubscript{0.88}     &   54.60\textsubscript{0.97}    &   56.45\textsubscript{0.97}   &  57.49   \\
\citet{jeon-strube-2020-incremental}\textsuperscript{${\dagger}$} & 57.30\textsubscript{\; \; \; \,}   &    61.70\textsubscript{\; \; \; \,}     &   54.50\textsubscript{\; \; \; \,}    &   56.90\textsubscript{\; \; \; \,}   &  57.60   \\ \hline \hline
XLNet+DNN & 60.70\textsubscript{1.03}   &    64.00\textsubscript{1.36}     &   55.15\textsubscript{1.14}    &   56.45\textsubscript{0.94}   &  59.10   \\  \hline \hline
Our Method & \textbf{63.65}\textsubscript{0.74}   &    \textbf{66.20}\textsubscript{0.81}     &   \textbf{57.00}\textsubscript{0.81}    &   \textbf{58.05}\textsubscript{1.21}   &  \textbf{61.23}   \\ \hline
\end{tabular}}
\caption{Mean accuracy (std) results on GCDC.}
\label{table-gcdc}
\vspace{-12pt}
\end{table*}

\noindent \textbf{Assessing Discourse Coherence}. ADC is the task of measuring the coherence of a given text. The benchmark dataset for this task is the Grammarly Corpus of Discourse Coherence (GCDC) dataset ~\citep{lai-tetreault-2018-discourse}. Specifically, GCDC contains texts from four domains, including \textbf{Yahoo} online forum posts, emails from Hillary \textbf{Clinton}'s office, emails from \textbf{Enron}, and \textbf{Yelp} online business reviews. It is annotated by expert raters with a coherence score $\in \{1, 2, 3\}$, representing low, medium, and high levels of coherence, respectively. 

\noindent \textbf{Automated Essay Scoring}. AES is a task to assign scores for essays, which has been used to evaluate coherence models ~\citep{burstein-etal-2010-using,jeon-strube-2020-incremental}. We follow previous work ~\citep{jeon-strube-2020-incremental} to employ the Test of English as a Foreign Language (TOEFL) dataset ~\citep{est2014} in our experiments. The corpus contains essays from \textbf{eight prompts} along with score levels (low/medium/high) for each essay.

\subsection{Experimental Settings}
We implement our method based on the Pytorch library. The pre-trained embedding we use to calculate the similarity between nouns is GloVe ~\citep{pennington-etal-2014-glove}, and we set the similarity threshold $\delta$ to 0.65. For the subgraph set construction, we use 4-node subgraphs as basic units for the ADC task and 5-node subgraphs for the AES task, and limit the maximum sentence distance $w$ to 8 for both tasks. The two-layer GCN is employed in our method, with ${\rm ReLU}$ as the activation function. We follow previous work ~\citep{jeon-strube-2020-incremental} to use the representation from ${\rm XLNet}_{base}$ as document node features, and initialize XLNet using the pre-trained checkpoint from Huggingface. We use XLNet instead of other pre-trained models, such as BERT, because the TOEFL dataset contains long texts. For example, some essays have more than 800 words (maybe more than 1000 subwords). Autoencoding-based pre-trained models, such as BERT, limit input text length (usually 512 subwords), whereas XLNet can handle any input sequence length. 

For the GCDC dataset, we follow the setting in ~\citet{lai-tetreault-2018-discourse} to perform 10-fold cross-validation over the training dataset. As for the TOEFL corpus, we conduct 5-fold cross-validation on the dataset of each prompt, which is a common setting for the AES task ~\citep{taghipour-ng-2016-neural}. Consistent with previous work ~\citep{lai-tetreault-2018-discourse,jeon-strube-2020-incremental}, we use mean accuracy (\%) as the evaluation metric. For more detailed settings and hyperparameters, please refer to Appendix \ref{sec:des_param}.

\noindent \textbf{Baselines}. To investigate the effectiveness of structural similarities between documents for coherence modeling, we empirically compare our method with a baseline model that does not use this knowledge. We call this baseline XLNet+DNN, which inputs document representations from XLNet as features, learns document embeddings with a two-layer deep neural network (DNN), and uses a softmax layer as the classifier. The only difference between the XLNet+DNN baseline and our method in terms of mathematical form is whether the regularized adjacency matrix $\tilde{D}^{-\frac{1}{2}}\tilde{A}\tilde{D}^{-\frac{1}{2}}$ is applied ~\citep{li2018deeper}. We configure this baseline to have the same number of parameters as our method for a fair comparison.

We also compare with ~\citet{mesgar-strube-2016-lexical}, which feeds subgraphs as input features. For a fair comparison, we input document representations from XLNet to this model, equip it with a two-layer DNN and softmax layer for feature extraction and classification. Furthermore, we compare our method against existing state-of-the-art models for each task to evaluate the effectiveness of our approach.

\begin{table*}[t]
\centering
\Large
\setlength{\tabcolsep}{3.2mm}
\scalebox{0.59}{
\begin{tabular}{l|cccccccc|c}
\hline
\multirow{2}{*}{Model} & \multicolumn{8}{l|}{\hspace{16.2em}Prompt}    &  \\
                       & 1\hspace{1.25em} & 2\hspace{1.25em} & 3\hspace{1.25em} & 4\hspace{1.25em} & 5\hspace{1.25em} & 6\hspace{1.25em} & 7\hspace{1.25em} & 8\hspace{1.25em} & Avg                     \\ \hline
\citet{dong-etal-2017-attention}    & 69.30\textsubscript{\; \; \; \,}  & 66.47\textsubscript{\; \; \; \,}  & 65.84\textsubscript{\; \; \; \,}  &  66.38\textsubscript{\; \; \; \,} & 68.89\textsubscript{\; \; \; \,}  &  64.20\textsubscript{\; \; \; \,} & 67.11\textsubscript{\; \; \; \,}  & 65.73\textsubscript{\; \; \; \,}  &   66.74    \\
\citet{mesgar-strube-2016-lexical}\textsuperscript{${\dagger}$}  & 75.31\textsubscript{0.77}  & 74.90\textsubscript{0.94}  & 73.42\textsubscript{0.81}  &  74.35\textsubscript{1.18} & 76.10\textsubscript{0.74}  &  75.42\textsubscript{0.68} & 72.48\textsubscript{0.83}  & 72.31\textsubscript{0.65}  &   74.29    \\
\citet{moon-etal-2019-unified}\textsuperscript{${\dagger}$} & 73.84\textsubscript{0.81}  & 72.54\textsubscript{0.87}  & 72.32\textsubscript{1.27}  &  73.26\textsubscript{0.67} & 75.34\textsubscript{0.72}  &  74.72\textsubscript{0.78} & 71.97\textsubscript{0.71}  & 72.14\textsubscript{0.93}  &   73.27    \\ 
\citet{jeon-strube-2020-centering}\textsuperscript{${\dagger}$} & 75.10\textsubscript{0.74}  & 73.35\textsubscript{0.92}  & 74.75\textsubscript{0.61}  &  74.18\textsubscript{1.07} & 76.38\textsubscript{0.91}  &  74.30\textsubscript{1.13} & 73.61\textsubscript{0.72}  & 73.44\textsubscript{1.15}  &   74.39    \\ 
\citet{jeon-strube-2020-incremental}\textsuperscript{${\dagger}$} & 75.60\textsubscript{\; \; \; \,}  & 73.40\textsubscript{\; \; \; \,}  & 75.00\textsubscript{\; \; \; \,}  &  73.50\textsubscript{\; \; \; \,} & 76.80\textsubscript{\; \; \; \,}  &  75.20\textsubscript{\; \; \; \,} & 73.50\textsubscript{\; \; \; \,}  & 72.80\textsubscript{\; \; \; \,}  &   74.48    \\ \hline \hline
XLNet+DNN  & 74.70\textsubscript{0.88}  & 74.46\textsubscript{0.97}  & 73.07\textsubscript{0.92}  &  74.09\textsubscript{1.04} & 75.45\textsubscript{0.83}  &  75.21\textsubscript{0.94} & 71.17\textsubscript{0.76}  & 71.95\textsubscript{0.81}  &   73.84    \\ \hline \hline    
Our Method  & \textbf{75.97}\textsubscript{1.14}  & \textbf{76.25}\textsubscript{1.07}  & \textbf{74.14}\textsubscript{1.18}  &  \textbf{75.81}\textsubscript{0.71} & \textbf{77.01}\textsubscript{0.94}  &  \textbf{77.08}\textsubscript{1.14} & \textbf{73.55}\textsubscript{0.80}  & \textbf{72.91}\textsubscript{0.66}  &   \textbf{75.34}    \\ \hline          
\end{tabular}
}
\caption{Mean accuracy (std) results on TOEFL.}
\label{table-toefl}
\vspace{-10pt}
\end{table*}

\subsection{Overall Results}

\noindent \textbf{Assessing Discourse Coherence}.
Table \ref{table-gcdc} shows the experimental results on GCDC dataset\footnote{In Tables \ref{table-gcdc}, \ref{table-toefl}, ${\dagger}$ denotes that the same XLNet as our method is employed in the model.}. The first three rows ~\citep{li-jurafsky-2017-neural,mesgar-strube-2018-neural,lai-tetreault-2018-discourse} in the first block show the performance of embedding-based models, and the last four rows \citep{mesgar-strube-2016-lexical, moon-etal-2019-unified,jeon-strube-2020-centering,jeon-strube-2020-incremental} in the same block are the state-of-the-art models based on XLNet. With the pre-trained model as the encoder, the latter four models outperform embedding-based methods by a large margin. 

We present the performance of the XLNet+DNN baseline and our method in the last two blocks of Table \ref{table-gcdc}. As shown in the table, structural similarity information between documents is helpful for coherence assessment, which improves the average accuracy from 59.10\% of the XLNet+DNN baseline to 61.23\% of our method. Subgraphs as input features ~\citep{mesgar-strube-2016-lexical} can also enhance performance, but the improvement is much smaller than our method. We speculate that simply concatenating subgraph features cannot efficiently capture structural similarities between documents. By contrast, our method explicitly connects structurally similar documents via a graph, thereby fully utilizing this information. Surprisingly, our simple baseline outperforms previous state-of-the-art models, which are also built on XLNet. This is likely because the GCDC dataset mainly contains short and informal texts, whereas previous sota models were designed to handle long and well-formed documents. By contrast, our method works well on the corpus, achieving the best performance. 

\noindent \textbf{Automated Essay Scoring}. As mentioned in Section \ref{sec-dataset}, AES is a task for scoring the quality of essays and has been used to evaluate coherence models. Hence, to better illustrate the effectiveness of our approach, we report the performance of both existing coherence models ~\citep{mesgar-strube-2018-neural,moon-etal-2019-unified,jeon-strube-2020-centering,jeon-strube-2020-incremental} and models designed to solve the AES task. For the latter, we report the result of ~\citet{dong-etal-2017-attention}, which is a state-of-the-art method for the AES task.

Results on the TOEFL dataset are shown in Table \ref{table-toefl}. Previous coherence models and the XLNet+DNN baseline give significantly better performance than the AES model in ~\citet{dong-etal-2017-attention}. Similar to the results on the GCDC dataset, subgraphs as input features can slightly improve the performance. However, the XLNet+DNN baseline can not beat the state-of-the-art coherence models on the TOEFL dataset. The results are reasonable because those coherence models are not only based on XLNet but also consider the characteristics of long documents. Consistent with observations on GCDC, our method, considering the structural similarities between documents, outperforms the XLNet+DNN baseline on the TOEFL dataset, giving state-of-the-art results. 

\begin{figure}[t]
\centering
\includegraphics[scale=0.48,trim=0 0 0 0]{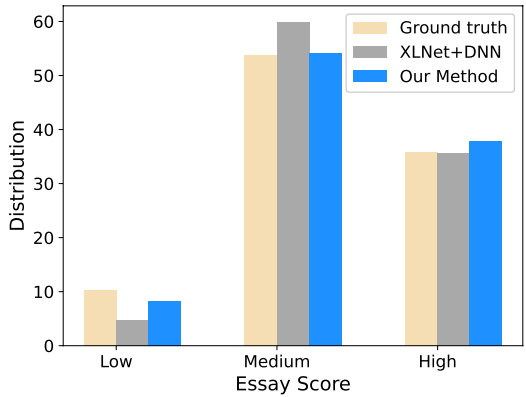}
\setlength{\abovecaptionskip}{5pt} 
\caption{Predicted label distribution.}
\label{fig-pred-bias}
\vspace{-8pt}
\end{figure}

\subsection{Performance Analysis}
\label{per_ana}
To understand how structural similarity works for coherence modeling, we compare our model with the XLNet+DNN baseline in terms of the predicted label distribution and document length.

\noindent\textbf{Predicted Label Distribution}. Figure \ref{fig-pred-bias} shows the distributions of predicted essay scores from the XLNet+DNN baseline and our model on the TOEFL P1 dataset. The XLNet+DNN's predictions are strongly biased, with about 60\% of essays predicted as medium scores. We speculate this is caused by the uneven label distribution in the TOEFL P1 dataset (10.3\%/53.8\%/35.9\% of low/medium/high-scoring essays). By contrast, our model is less affected by the uneven distribution, making more low and high score predictions. We also collect the prediction accuracy of the two models for each essay score. The prediction accuracy of the XLNet+DNN model for low, medium, high scores is 35.29\%, 83.71\%, 76.47\%, and that of our method is 50.00\%, 82.02\%, 84.87\%. XLNet+DNN mainly predicts medium scores, so this label's recall value is high. Compared with the baseline, our method makes relatively accurate predictions for all essay scores, suggesting that capturing structural similarities between essays helps mitigate the effects of uneven label distribution and thus focuses on learning coherence patterns.

\begin{table}[t]
\Large
\centering
\scalebox{0.56}{
\begin{tabular}{l|llll|c}
\hline
Model      & Yahoo & Clinton & Enron & Yelp  & Avg   \\ \hline
XLNet+DNN  & 47.32\textsubscript{1.56} & 46.16\textsubscript{1.77}   & 42.86\textsubscript{1.85} & 39.32\textsubscript{1.73} & 43.91 \\ \hline
Our Method & \textbf{51.92}\textsubscript{1.06} & \textbf{48.49}\textsubscript{1.61}   & \textbf{45.67}\textsubscript{1.57} & \textbf{44.18}\textsubscript{1.10} & \textbf{47.66} \\ \hline
\end{tabular}
}
\caption{Mean F1-Macro results (std) on the GCDC.}
\label{f1-gcdc}
\vspace{-12pt}
\end{table}

\begin{figure*}[t]
\centering\includegraphics[scale=0.46,trim=0 0 0 0]{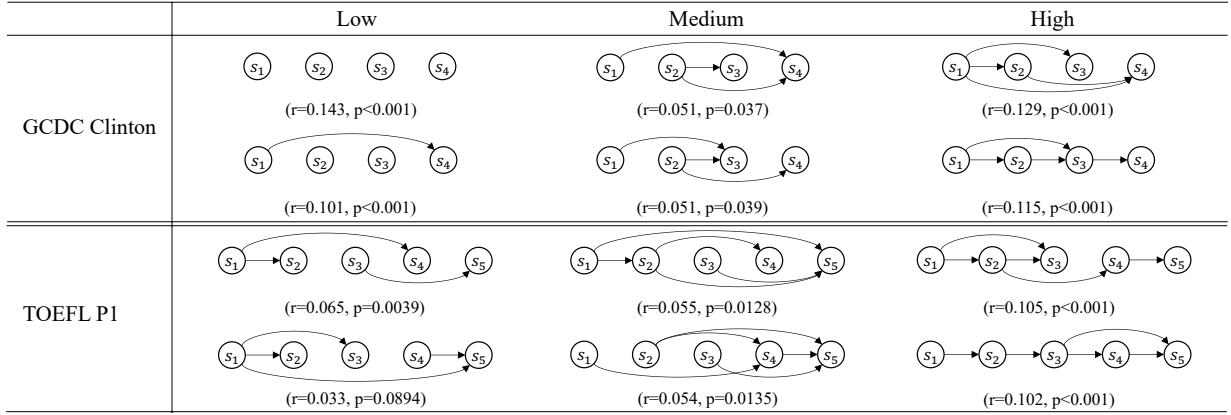}
\setlength{\abovecaptionskip}{-8pt} 
\caption{The top two most positively correlated subgraphs for each coherence level on the GCDC Clinton and TOEFL P1. $r$ denotes the correlation coefficient value, and $p$ is the p\_value ($p$ < 0.05 means statistically significant).}
\label{top-subgraph}
\vspace{-6pt}
\end{figure*}

To better verify this, we report the performance of the XLNet+DNN baseline and our method using F1-Macro as the evaluation metric. F1-Macro computes the accuracy for each class independently and then takes the average at the class level. Intuitively, if our model’s predictions are more uniform and accurate, the F1-Macro performance gap between our method and the XLNet+DNN baseline should be no smaller than the gap in terms of accuracy. Table \ref{f1-gcdc} shows the F1-Macro results of the XLNet+DNN baseline and our model on the GCDC dataset. Our method achieves much better F1-Macro results than the XLNet+DNN baseline, and the gap between the two models in F1-Macro is larger than the gap in accuracy, which further demonstrates that our model makes more even and accurate predictions.

\begin{figure}[t]
\centering
\includegraphics[scale=0.45,trim=0 0 0 0]{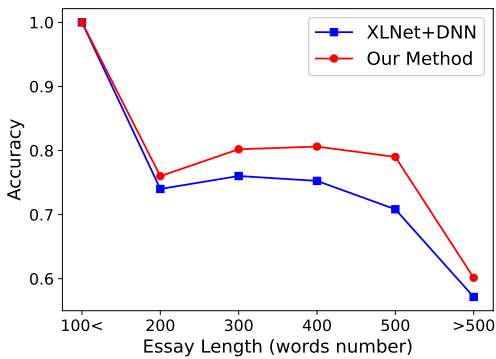}
\setlength{\abovecaptionskip}{5pt} 
\caption{Accuracy against essay length. }
\label{fig-doc-length}
\vspace{-8pt}
\end{figure}

\noindent\textbf{Document Length}. Figure \ref{fig-doc-length} shows the accuracy trends of the baseline and our method on the TOEFL P1 dataset as essays become longer. The curve of XLNet+DNN generally shows a downward trend, accuracy decreasing as essays' length increases. The result is not surprising since long documents contain more complicated semantics and thus are more challenging. Our method performs similarly to the XLNet+DNN baseline over short documents (length <= 200). But when essays become longer, our model gives relatively high accuracy and even presents a slight increase (200 < length <= 400). This suggests that structural similarity information helps to improve the model's robustness when document length increases.

\subsection{Ablation Study}
\label{alb_sec}
We analyze the effectiveness of each type of edge in our method. To this end, we test the performance of our approach by first removing edges between subgraph nodes (ESS) and then removing edges between document node and subgraph node (EDS). Note that if all edges are removed (i.e. each document is an isolated node), our method degrades into the XLNet+DNN baseline. 

Table \ref{ablation} shows the results on the GCDC Clinton and TOEFL P1 datasets. We can observe from the table that eliminating any type of edges would hurt the performance. The decline in performance is more significant when removing the EDS than eliminating the ESS. The results are reasonable because edges between documents and subgraphs are the key to connecting documents with similar structures, while edges between subgraphs are considered to further assist it ~\citep{kondor2009graphlet}.

\begin{table}[t]
\centering
\scalebox{0.84}{
\begin{tabular}{l|cc}
\hline
\multirow{2}{*}{Model} & \hspace{1em}GCDC Clinton\hspace{1em}   & \hspace{1em}TOEFL P1\hspace{1em}  \\ \cline{2-3}
                       & Acc & Acc \\ \hline
Our Method             &  \textbf{66.20}         &    \textbf{75.97}       \\ 
- ESS                  &    66.00       &   75.42        \\ 
- ESS, EDS             &   64.00        & 74.70 \\ \hline             
\end{tabular}}
\caption{Ablation study for different edges on the GCDC Clinton and TOEFL P1 dataset.}
\label{ablation}
\vspace{-12pt}
\end{table}

\subsection{Subgraph Analysis}
\label{sec-sub-study}
In this section, we statistically investigate which subgraphs (sentence connection styles) mostly correlate to each level of coherence\footnote{We perform this analysis on the whole corpus and show readable text examples in Appendix \ref{subg-example-sec}}. Specifically, we calculate the Pearson correlation coefficient between each subgraph and per label, and test the significance of the correlation. Figure \ref{top-subgraph} shows the top two results on GCDC Clinton and TOEFL P1. 

In general, subgraphs positively correlated with higher coherence tend to contain more edges. This is somewhat aligned with the previous finding \citep{guinaudeau-strube-2013-graph} that coherence correlates with the average out-degree of sentence graphs. Weakly connected subgraphs are more likely to reflect higher coherence than disconnected ones. Taking results on GCDC Clinton as an example, the top two most correlated subgraphs for low coherence contain isolated nodes or components while nodes in subgraphs for high coherence are (weakly) connected. Furthermore, subgraphs with more connections between adjacent sentences seem more correlated with high coherence. For example, there is an almost linear subgraph (or contains linear structure) in the high category of both datasets. We also find that the subgraph results per coherence level on the GCDC Clinton dataset differ from that on the TOEFL P1 dataset. This could be due to two reasons. First, the two datasets contain texts from various domains with domain-specific writing styles and structures. Second, they are built by different annotators, who may have different preferences for text organization styles.

\section{Conclusion}
In this paper, we investigated the effectiveness of structural similarity information between documents for coherence modeling. We proposed a graph-based method to connect structurally similar documents based on shared subgraphs, and model the connectivity relations with a GCN. Experiments on two benchmark tasks show that our method consistently outperforms the baseline model, achieving state-of-the-art results on both tasks.

\section{Limitations}
Despite showing impressive performance, our graph-based approach still has several limitations. The first one is related to the construction of the sentence graph. At present, we consider two sentences to be semantically related if they share similar nouns. But coherence can be achieved not only by describing similar entities but also by discourse (rhetorical) relations ~\citep{speech}. So it will be an exciting direction to incorporate discourse relations into the construction of a graph. The second one is that we implemented our method using only a plain GCN. Recent work has pointed out that the original GCN can be further improved with more advanced aggregation functions \citep{XuHLJ19} or attention mechanisms \citep{VelickovicCCRLB18}. So another interesting direction is to explore the benefits of more powerful graph neural networks for our method, which we leave for future study.


\bibliography{main.bbl}
\bibliographystyle{acl_natbib}

\clearpage

\appendix
\section{Graph Construction}
\label{sec:al_graph}
\vspace{-5pt}
\begin{algorithm}[h]
	\renewcommand{\algorithmicrequire}{\textbf{Input:}}
	\renewcommand{\algorithmicensure}{\textbf{Output:}}
	\caption{Constructing sentence graph}
	\label{alg_sentgraph}
	\begin{algorithmic}[1]
	    \Require
	        Document $d$, threshold $\delta$
	    \Ensure
	        Sentence graph $\rm G$
	    \State $S, NS$ $\gets$ stanza($d$) \Comment{Sentences and nouns}
		\State $L$ $\gets$ len($S$) 
		\State $G$ $\gets$ zeros($L$, $L$)  \Comment{Init adjacency matrix}
		\For{$u \gets 1$ \textbf{to} $L-1$} 
		    \For{$v \gets u+1$ \textbf{to} $L$} 
		        \State $un, vn \gets $ len($NS_u$), len($NS_v$)
	
		        \State $sim\_scores \gets [$\;$] $
		        
		        \For{$a \gets 1$ \textbf{to} $un$}
		            \For{$b \gets 1$ \textbf{to} $vn$}
		                \State $e_a \gets $ embed($NS_{u,a}$)
		                \State $e_b \gets $ embed($NS_{v,b}$)
		                \State $score \gets$ cos\_sim($e_a, e_b$)
		                \State Append($score$, $sim\_scores$)
		            \EndFor
		        \EndFor
		        \State $max\_score \gets$ max($sim\_scores$)
		        \If{$max\_score > \delta $}
		            \State $G_{u,v} \gets 1 $
		        \EndIf
		    \EndFor
		\EndFor
	\end{algorithmic}  
\end{algorithm}

\begin{algorithm}[H]
	\renewcommand{\algorithmicrequire}{\textbf{Input:}}
	\renewcommand{\algorithmicensure}{\textbf{Output:}}
	\caption{Counting Subgraph Frequency}
	\label{alg_subgraph}
	\begin{algorithmic}[1]
	    \Require
	        Sentence graph $G$, subgraph size $k$, max sentence distance $w$
	    \Ensure
	        subgraph set $freq$
	    \State $freq \gets \{\}$ \Comment{frequency of each subgraph}
	    \State $nodes \gets G.nodes()$
        \State $i,\; n \gets 0,\;$ len($nodes$)
        \While{$i < (n-k+1)$}
            \State $w\_n \gets nodes[i:i+w]$ \Comment{distance $< w$}
            \State $k\_node\_combs \gets$ combinations($w\_n$, $k$)
            \For{$k\_nodes$ \textbf{in} $k\_node\_combs$}
            \State $subgraph \gets $ extract($G$, $k\_nodes$)
            \State $signature \gets $ pynauty($subgraph$)
            \State Add($freq[signature]$, 1)
        \EndFor
        \State $i \gets i + (w - k + 1)$
        \EndWhile
	\end{algorithmic}  
\end{algorithm}

\vspace{-10pt}
\begin{algorithm}[h]
	\renewcommand{\algorithmicrequire}{\textbf{Input:}}
	\renewcommand{\algorithmicensure}{\textbf{Output:}}
	\caption{Evaluation}
	\label{alg_eval}
	\begin{algorithmic}[1]
	    \Require
	        Test corpus $\textbf{TC}$, Doc-subgraph graph $G$, Trained $\rm GCN$
	    \Ensure
	        Predictions $preds$
	    \State $preds \gets$ []
		\State $N$ $\gets$ len($\textbf{TC}$) 
		\For{$i \gets 1$ \textbf{to} $N$} 
			\State $d_i \gets \textbf{TC}[i]$
			\State $G^{*} \gets$ Add($d_i$, $G$) \Comment{Add document}
		    \State $G^* \gets$ Norm($G^*$) \Comment{Norm graph\;\;\;\;}
		    \State $l_i \gets$ GCN($G^*$) \Comment{Predict label\;\;\;}
		    \State Append($l_i$, $preds$)
		\EndFor
	\end{algorithmic}  
	\label{al:eval}
\end{algorithm}
\vspace{-5pt}

\begin{table}[t]
\centering
\Large
\scalebox{0.56}{
\begin{tabular}{|ll|l|c|c|c|c|}
\hline
\multicolumn{2}{|l|}{Dataset}                                           & Split & \#Doc & Avg \#W & Max \#W & Avg \#S \\ \hline
\multicolumn{1}{|l|}{\multirow{8}{*}{GCDC}}  & \multirow{2}{*}{Yahoo}   & Train & 1000  & 157.2         & 339    &  7.8   \\ \cline{3-7} 
\multicolumn{1}{|l|}{}                       &                          & Test  & 200   & 162.7         & 314    &  7.8  \\ \cline{2-7} 
\multicolumn{1}{|l|}{}                       & \multirow{2}{*}{Clinton} & Train & 1000  & 182.9         & 346    &  8.9  \\ \cline{3-7} 
\multicolumn{1}{|l|}{}                       &                          & Test  & 200   & 186.0         & 352    &  8.8  \\ \cline{2-7} 
\multicolumn{1}{|l|}{}                       & \multirow{2}{*}{Enron}   & Train & 1000  & 185.1         & 353    &  9.2  \\ \cline{3-7} 
\multicolumn{1}{|l|}{}                       &                          & Test  & 200   & 191.1         & 348    &  9.3  \\ \cline{2-7} 
\multicolumn{1}{|l|}{}                       & \multirow{2}{*}{Yelp}    & Train & 1000  & 178.2         & 347    &  10.4  \\ \cline{3-7} 
\multicolumn{1}{|l|}{}                       &                          & Test  & 200   & 179.1         & 340    & 10.1   \\ \hline
\multicolumn{1}{|l|}{\multirow{8}{*}{TOEFL}} & Prompt 1                 & Total & 1656  & 339.1         & 806    &  13.7  \\ \cline{2-7} 
\multicolumn{1}{|l|}{}                       & Prompt 2                 & Total & 1562  & 357.8         & 770    &  15.7  \\ \cline{2-7} 
\multicolumn{1}{|l|}{}                       & Prompt 3                  & Total & 1396  & 343.5         & 731   &  14.7   \\ \cline{2-7} 
\multicolumn{1}{|l|}{}                       & Prompt 4                  & Total & 1509  & 338.0         & 699   &  15.1   \\ \cline{2-7} 
\multicolumn{1}{|l|}{}                       & Prompt 5                 & Total & 1648  & 358.4         & 876    & 15.2   \\ \cline{2-7} 
\multicolumn{1}{|l|}{}                       & Prompt 6                 & Total & 960         & 358.3    & 784   & 15.3  \\ \cline{2-7} 
\multicolumn{1}{|l|}{}                       & Prompt 7                 & Total & 1686  & 336.6         & 638    & 14.0   \\ \cline{2-7} 
\multicolumn{1}{|l|}{}                       & Prompt 8                 & Total & 1683  & 340.9         & 659    & 14.7   \\ \hline
\end{tabular}
}
\caption{The statistics of datasets. \#Doc, \#W, \#S denotes the number of documents, words, sentences.}
\label{table-stat}
\vspace{-10pt}
\end{table}

\section{Train and Evaluation}
\label{sec:train_and_eval}
Vanilla GCN is a transductive method in which both training and test data are presented to the model during training. This, however, is not applicable in practice since we do not know the evaluation documents in advance. To overcome this drawback, we implement an inductive GCN inspired by the work in fast GCN ~\citep{chen2018fastgcn}. Specifically, we first construct the doc-subgraph graph based on the training corpus (Section \ref{doc-sub-graph}) and train GCN on this graph (Section \ref{gcn}). While evaluating, for each document in the test corpus, we add it to the doc-subgraph graph, normalize the adjacency matrix of the new graph, and predict its label (refer to Algorithm \ref{al:eval}). Consequently, our method is in a pure inductive setting. That is, our model does not see the test corpus during training, and its evaluation is performed on individual documents without using the information of other samples in the test corpus. Note that when calculating weights for edges between the newly added document node and subgraph nodes, the inverse document frequency we used in equation (\ref{tf-idf}) is the one we computed using only the training corpus.

\section{Dataset Description}
\label{sec:dd_dataset}
The statistics of the GCDC and TOEFL datasets is shown in Table \ref{table-stat}. 

\begin{figure*}[t]
\centering\includegraphics[scale=0.46,trim=0 0 0 0]{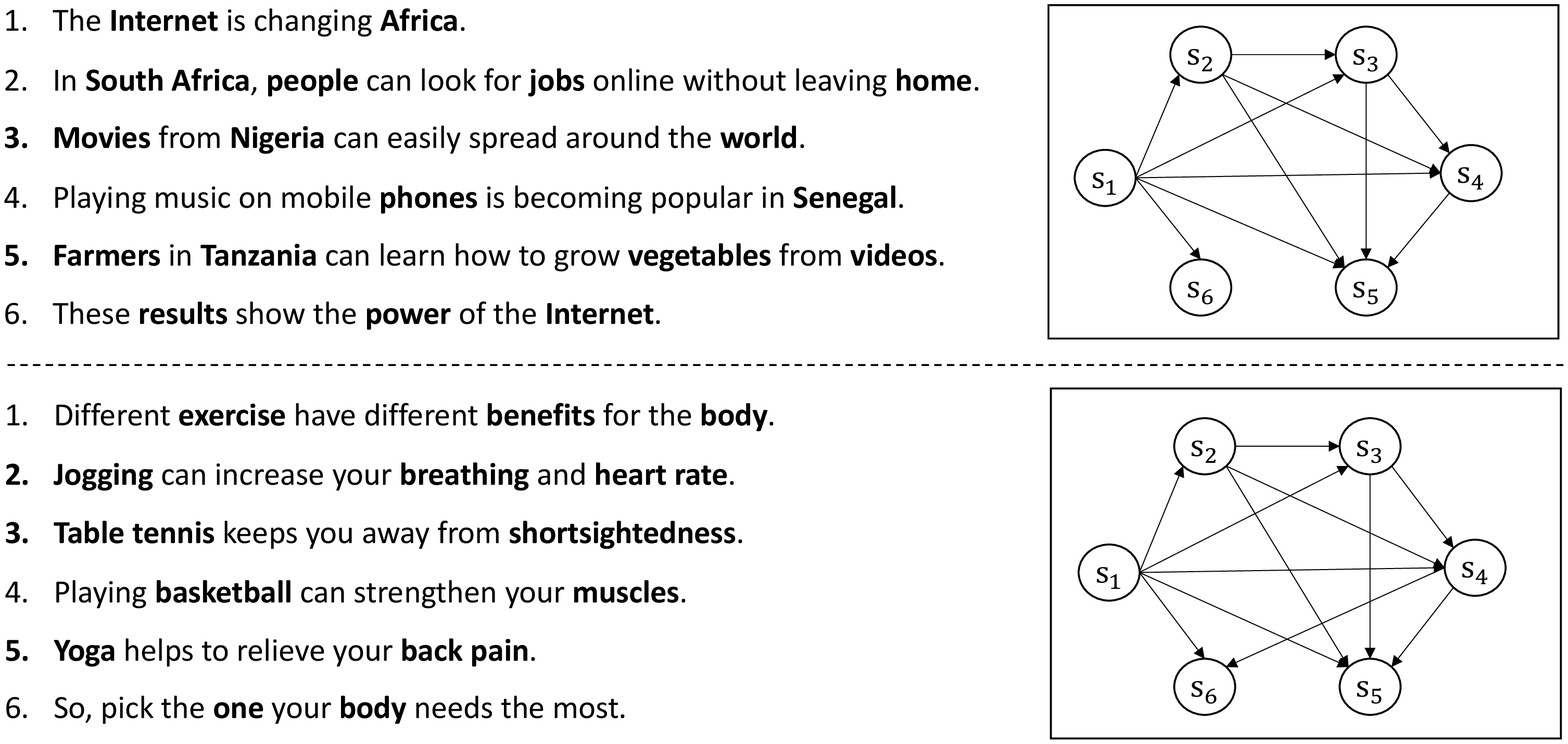}
\setlength{\abovecaptionskip}{8pt} 
\caption{An example of two highly coherent texts with similar connection structures. We bold the recognized nouns in the example.}
\label{fig-compare}
\vspace{-12pt}
\end{figure*}

\section{Detailed Experimental Setting}
\label{sec:des_param}
For the GCDC dataset, we perform 10-fold cross-validation over the training dataset following previous work ~\citep{lai-tetreault-2018-discourse}. The dimensionality of the two-layer GCN is set to be 240 for Clinton and Enron domains, and 360 for Yahoo and Yelp domains. We use the Adam optimizer with an initial learning rate of 0.01 on Clinton and Enron, 0.008 on Yahoo and Yelp. As for the TOEFL corpus, we conduct 5-fold cross-validation on the dataset of each prompt, which is the common evaluation setting for the AES task ~\citep{taghipour-ng-2016-neural}. A two-layer GCN with dimension size 240 and the Adam optimizer with an initial learning rate of 0.05 is employed for every prompt dataset. Dropout with a rate of 0.5 is applied to both tasks. And we train the model for 160 epochs on the GCDC dataset and 400 on the TOEFL dataset. For the XLNet+DNN baseline, we configure it with the same trainable parameters as our method. As for ~\citet{mesgar-strube-2016-lexical}, we concatenate the document presentation from XLNet and subgraphs mining from the document's sentence graph as input, and also use a two-layer DNN and a softmax layer. Note that this model has more trainable parameters since its input dimension becomes larger (the concatenation of subgraphs and XLNet representation). For other baseline models, we apply the same experimental setting and XLNet to them as our method and tune their hyperparameters according to the performance on the Dev set. We conduct all experiments on a single Tesla P40 GPU with 24GB memory. It takes about 0.5 days to train our model on the GCDC dataset and 1.5 days on the TOEFL dataset.

We follow previous works ~\citep{lai-tetreault-2018-discourse, taghipour-ng-2016-neural} to use the mean of multi-run accuracy (std) as the evaluation metric. 

\section{Examples}
\subsection{Text Example}
\label{app.sub.text_exam}
Coherence describes how sentences of a text connect to each other ~\citep{tanya1980,foltz1998measurement,schwarz2001establishing}. Theoretically, documents with similar connection structures should tend to have similar degrees of coherence. To help readers understand it, we show two texts in Figure \ref{fig-compare}. Although the two texts have different content, they share very similar connection structures. For example, the first text first talks about Africa, then discusses specific African countries, and finally makes a conclusion. The second text starts with exercise, then goes to certain daily sports, and finally makes a summary. Based on the linguistic definition of text coherence, the two texts should have a similar degree of coherence due to their similar organizational structures. This could be a very useful prior knowledge when we measure a text's coherence. For example, in Figure \ref{fig-compare}, we can easily assess the coherence of one text by referring to the label of the other one since they have very similar organizational structures.

\begin{figure*}[t]
\centering\includegraphics[scale=0.34,trim=0 0 0 0]{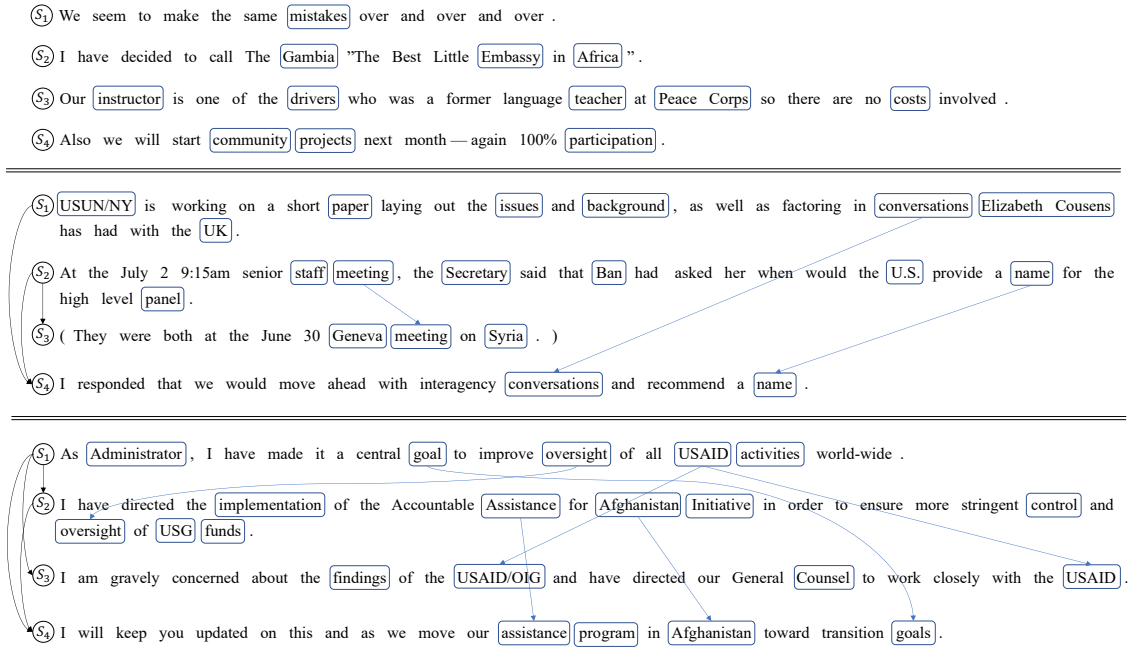}
\setlength{\abovecaptionskip}{1pt} 
\caption{Three text examples with constructed subgraphs from the GCDC Clinton dataset. We show subgraphs of each text example to the left of that example.}
\label{clinton-example}
\end{figure*}

\begin{figure*}[t!]
\centering\includegraphics[scale=0.34,trim=0 0 0 0]{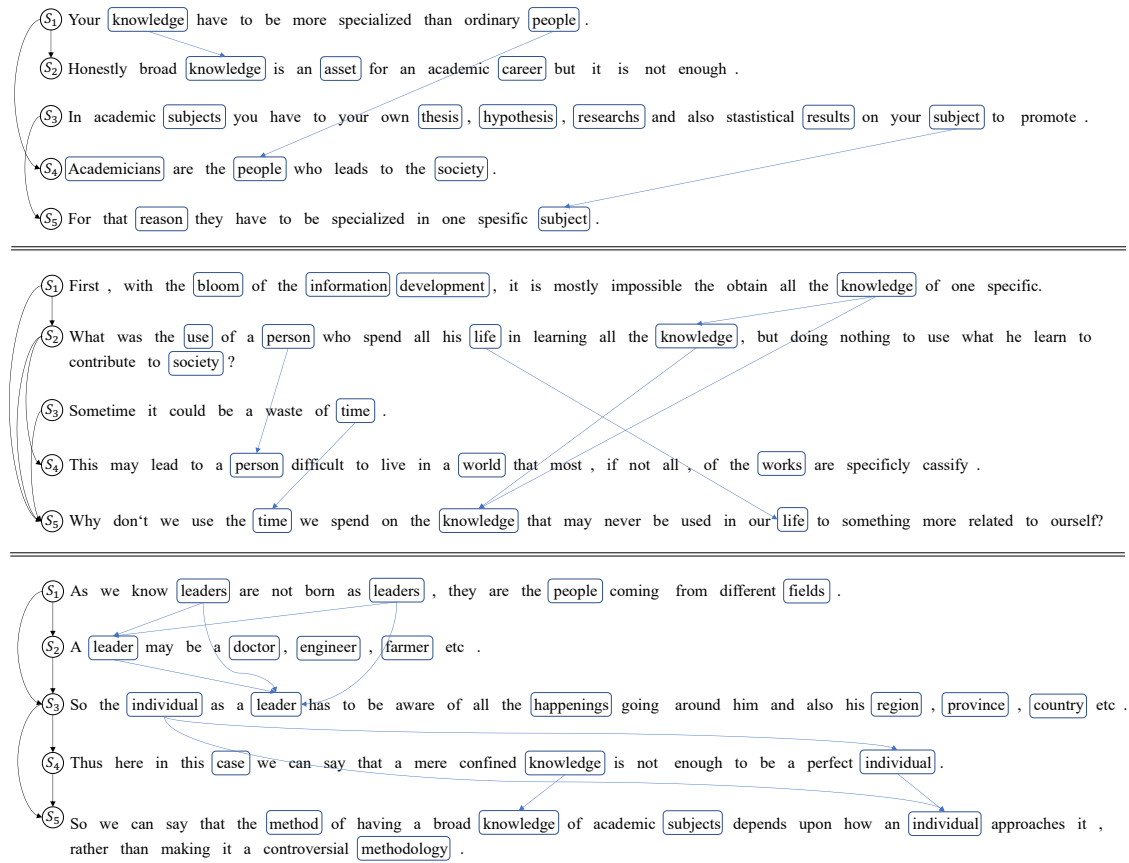}
\setlength{\abovecaptionskip}{1pt} 
\caption{Three text examples with constructed subgraphs from the TOEFL Clinton dataset.}
\label{toefl-example}
\vspace{-12pt}
\end{figure*}

\subsection{Subgraph Examples}
\label{subg-example-sec}
We show several text pieces with constructed subgraphs in Figure \ref{clinton-example} (from the GCDC Clinton dataset) and Figure \ref{toefl-example} (from the TOEFL P1 dataset). In each example, the corresponding subgraph is shown on the left. We use blue boxes to mark the recognized nouns in each sentence and link semantically related nouns between different sentences by a directed edge between two boxes. Two sentences will be connected if there are semantically related nouns between them.

\end{document}